\pdfoutput=1

\documentclass[11pt]{article}

\usepackage[final]{acl}

\usepackage{times}
\usepackage{latexsym}
\usepackage{amsmath}

\usepackage{tabularx}
\usepackage{url}
\usepackage{graphicx}
\usepackage{algorithm}
\usepackage{algorithmic}
\usepackage{booktabs}
\usepackage{multirow} 
\usepackage{csquotes} 
\usepackage[T1]{fontenc} 
\usepackage{longtable}
\usepackage{xcolor}
\usepackage{arydshln}
\usepackage[most]{tcolorbox}

\newtcolorbox{mybox}[2][]{
  width=\textwidth,
  colback=white, 
  colframe=pastelblue,
  fonttitle=\bfseries,
  coltitle=white, 
  title=#2,
  #1,
  colbacktitle=pastelblue, 
  enhanced,
  attach boxed title to top left={yshift=-2mm, xshift=2mm},
  boxed title style={colframe=pastelblue},
  separator sign={\ ---\ }
}
\definecolor{pastelblue}{rgb}{0.85, 0.85, 1.0}
\usepackage[T1]{fontenc}

\usepackage[utf8]{inputenc}

\usepackage{microtype}

\usepackage{inconsolata}

\usepackage{graphicx}

\newcommand{\mymodel}{DVR}
\title{Divide-Verify-Refine: Can LLMs Self-Align with Complex Instructions?}

\author{Xianren Zhang\textsuperscript{1}, Xianfeng Tang\textsuperscript{2}, Hui Liu\textsuperscript{2}, Zongyu Wu\textsuperscript{1}, Qi He\textsuperscript{2}\\ \textbf{Dongwon Lee}\textsuperscript{1}, \textbf{Suhang Wang}\textsuperscript{1}\\
\textsuperscript{1}The Pennsylvania State University  \quad \textsuperscript{2}Amazon\\
\texttt{\{xzz5508,dongwon,szw494\}@psu.edu}, \texttt{\{liunhu,xianft\}@amazon.com}
}

\begin{document}
\maketitle
\begin{abstract}
Recent studies show LLMs struggle with complex instructions involving multiple constraints (e.g., length, format, sentiment). 
Existing works address this issue by fine-tuning, which heavily relies on fine-tuning data quality and is computational expensive.  
An alternative is leveraging LLMs' self-correction to refine responses for better constraint adherence. However, this is limited by the feedback quality, as LLMs cannot generate reliable feedback or detect errors. Moreover, its effectiveness relies on few-shot examples illustrating response modifications. As constraints in complex instructions are diverse, manually crafting such examples for each constraint type can be labor-intensive and sub-optimal. To address these two challenges, we propose the \textbf{Divide-Verify-Refine (DVR)} framework with three steps: (1) \textbf{Divide} complex instructions into single constraints and prepare appropriate tools; (2) \textbf{Verify} responses using tools that provide rigorous check and textual guidance (e.g., Python toolkit for format checks or pre-trained classifiers for content analysis); (3) \textbf{Refine}: To maximize refinement effectiveness, we propose dynamic few-shot prompting, where a refinement repository collects successful refinements, and these examples are selectively retrieved for future refinements. Recognizing the lack of complexity in existing datasets, we create a new dataset of complex instructions. DVR doubles Llama3.1-8B's constraint adherence and triples Mistral-7B's performance. The code is available  \href{https://anonymous.4open.science/r/CODE_Constraint_Following/README.md}{here}.
\end{abstract}

\section{Introduction}

Large language models (LLMs), like ChatGPT, have shown significant improvements across various language tasks \citep{touvron2023llama,wang2024comprehensive,zhang2024catastrophic}. The success of LLMs relies on the ability to execute complex instructions. Failures to follow instructions can result in unintended outputs, which may have severe consequences \citep{mu2023can,zhou2023instruction}. This issue becomes critical when LLMs are deployed in high-stakes environments, such as legal documentation or technical writing. For example, when drafting legal contracts, LLMs must strictly adhere to constraints related to format, specific terminology, and precise language usage to avoid misinterpretations or legal liabilities. Similarly, in technical writing, adhering to strict format guidelines, word limits, and inclusion of essential technical terms is critical to ensure clarity and compliance with industry standards.

Recent studies show that LLMs, especially open-source ones, struggle to follow complex instructions with multiple constraints like response length or formatting \citep{he2024complex,jiang2023followbench,chen2024benchmarking}. While this issue is well recognized, research on enhancing LLMs' constraint adherence ability is still limited, with most efforts focused on evaluating the ability \citep{jiang2023followbench,chen2024benchmarking}. Few studies improve LLMs' constraint-following via fine-tuning \citep{he2024complex,sun2024conifer,li2024ruler}. For example, \citeauthor{he2024complex}~\citeyear{he2024complex}  adopt a teacher model (e.g., GPT-4) to generate data and fine-tune a student model with the generated data to improve its multi-constraint adherence ability. While effective, it requires a large amount of computation resources and heavily depends on the generated data quality. 

\begin{figure}[t]
\includegraphics[width= 0.9\columnwidth]{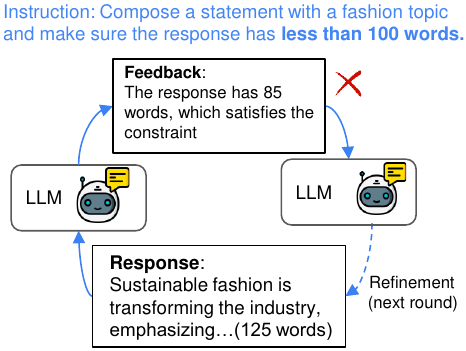}

\caption{The LLMs hallucinate and struggle to give reliable feedback.}
\label{fig:intro_fig}
\vskip -1em
\end{figure}

In contrast, the concept of ``self-correction'' offers an alternative approach, where LLMs autonomously correct their responses \citep{madaan2024self,shinn2024reflexion}. Self-correction has been applied on other tasks such as question answering \citep{dhuliawala2023chain, shinn2024reflexion} or mathematics \citep{madaan2024self}, where an LLM will evaluate its responses, give feedback, and further refine responses. However, {\em whether LLMs can effectively self-align with diverse and complex constraints remains an open question.} This is particularly important for agent LLMs to be deployed in high-stakes environments.
For constraint-following, this self-correction process can be divided into two phases: verification and self-refinement (Fig. \ref{fig:intro_fig}). During the verification phase, LLMs assess whether their responses align with the specified constraints. If the responses do not align with the constraints, the LLMs will give feedback that pinpoints errors and suggests adjustments. Following this, the self-refinement phase takes place where LLMs use the feedback to refine and improve their responses accordingly.


However, there are several challenges. The first one is {\em feedback reliability}. 
Studies show that the performance gains by LLM self-correction is unstable, occasionally even degrading performance in question answering \citep{huanglarge} and code generation \citep{olausson2023self}. 
The bottleneck in self-correction lies in the feedback quality \citep{tyen2024llms, goucritic, jiang2024self}. LLMs, including advanced models like GPT-4 and Claude 3, tend to have low recall in detecting LLMs errors, underperforming significantly compared to humans \citep{kamoi2024evaluating}. On the other hand, research reveals that the self-correction performance on reasoning tasks is boosted if the error location is given, indicating LLMs have the self-correction ability given reliable feedback \citep{tyen2024llms}. From the constraint-following perspective, LLMs are also not good at checking simple and easy-to-verify constraints. As shown in Fig.~\ref{fig:intro_fig}, given a response, LLMs struggle to accurately count the number of words. The second challenge is {\em constraint diversity} which lies in the self-refinement process. Given the response and the feedback, LLMs should refine the response according to the feedback. However, to perform this task effectively, a set of representative few-shot examples is needed to demonstrate how to appropriately modify the response \citep{brown2020language}. These constraints can vary widely, from adhering to a length limit to including specific keywords. Each type of constraint needs distinct modifications. For example, meeting a length limit might require removing content, whereas incorporating specific keywords requires adding text. Manually crafting representative few-shot examples for each constraint type is labor-intensive.

\begin{figure*}[t]
\centering
\includegraphics[width=\textwidth]{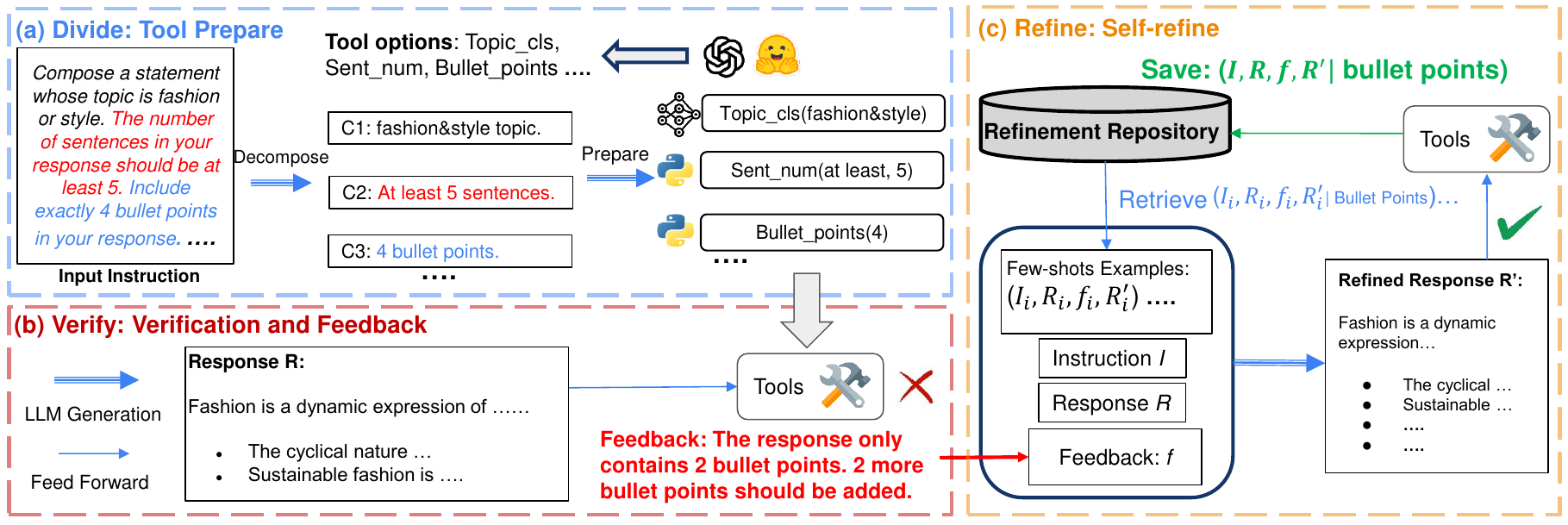}
\vskip -0.5em
\caption{The {\mymodel} framework: (a) Divide: The LLMs decompose constraints and instantiate tools for each constraint, (b) Verify: Tools will give feedback on the response, (c) Refine: The refinement repository provides past refinement process as few-shot examples. The current refinement process will be stored in the repository. 
}
\label{fig:framework}
\end{figure*}

To address these challenges, we propose a novel framework named Divide-Verify-Refine ({\mymodel})
as shown in Fig. \ref{fig:framework}. To enhance feedback reliability, we observe that constraints that LLMs struggle to verify can be readily assessed using external tools. These tools include python toolkit for quantitative measures, such as Regular Expressions (re) \citep{friedl2006mastering} and the Natural Language Toolkit (NLTK) \citep{bird2009natural} for counting the number of words, sentences, paragraphs, or bullet points, as well as pre-trained classifiers for content analysis, such as topic and sentiment analysis, which are easily accessible and widely available on hugging face \citep{antypas2022twitter,loureiro2022timelms}. We enhance LLMs by enabling interaction with external tools. First, we instruct LLMs to break down complex instructions into individual constraints and assign an appropriate tool to each. These tools then rigorously verify the LLM’s response and provide textual guidance for refinement if any constraints are violated. To address the problem of constraint diversity, we propose the {\em dynamic few-shot prompting}. Since the verification reliability is ensured by external tools, we incorporate a novel refinement repository, which serves as a memory module to collect and store successful refinements for future use. When a new refinement task arises, we select few-shot examples with the same constraint type to maximize refinement effectiveness.

Our \textbf{main contributions} are: (i) This is the first work to improve LLMs' constraint-following ability without training. Our framework enhances feedback reliability by integrating easily accessible tools that provide strict verification and textual guidance for LLMs. (ii) To maximize the refinement effectiveness, we propose {\em dynamic few-shot prompting} with a refinement repository that stores successful refinements. This enables LLMs to learn from past experiences and retrieve more similar few-shot examples for future refinements, improving refinement effectiveness. (iii) Most benchmarks only contain 1-2 constraints \citep{chen2024benchmarking}. We construct a new complex instruction dataset with instructions containing 1-6 constraints.

\section{Related Work}

\textbf{Instruction-Following of LLMs.} Recent studies show that LLMs struggle to follow complex instructions, especially as the number of constraints increases \citep{dubois2024length,zhou2023instruction,jiang2023followbench,chen2024benchmarking,zhou2023instruction,he2024can}. To address this challenge, some works \citep{chen2023comprehensive,sun2024conifer,wang2024instructions,he2024complex,dong2024self} generate instructions and responses with advanced LLMs (e.g., GPT4) and then use the generated data to fine-tune the student LLMs. Among them, \citet{he2024complex} focuses on improving LLMs' alignment with multiple constraints. They iteratively refine student model responses using GPT-4 as a teacher. The student model is fine-tuned on both intermediate modifications and final refined responses. Although effective, these methods rely heavily on the teacher model and are resource-consuming. Different from previous methods, our framework uses ins-context learning with tool interaction to effectively refine unsatisfactory responses, offering a more practical solution.

\textbf{Self-Correction of LLMs.} Self-correction is a framework where LLMs refine their responses during inference by reflecting on their initial responses \citep{shinn2024reflexion,madaan2024self}. This process has two phases. Initially, LLMs are prompted to analyze and provide feedback on their responses. Subsequently, based on the feedback LLMs refine the responses to correct their mistakes. However, recent studies report negative results indicating that LLMs cannot self-correct their own mistakes \citep{hong2024closer,tyen2024llms,kamoi2024evaluating,goucritic}. A study \cite{kamoi2024evaluating} reveals that top LLMs like GPT-4 and Claude 3 have low recall in detecting LLM errors, with LLMs significantly underperforming compared to humans. Additionally, feedbacks provided by LLM self-correction tend to hallucinate and lack reliability. This unreliability suggests that even when errors are detected, the guidance offered for corrections may be incorrect or misleading. \cite{hong2024closer} find that LLMs struggle to accurately identify logical fallacies, casting doubt on their inherent ability to detect errors and conduct self-verification reasoning effectively. However, the self-correction performance on reasoning tasks is boosted if the error location is given \citep{tyen2024llms}. All these observations indicate that LLMs are not reliable in analyzing their responses and a more reliable feedback mechanism is needed to pinpoint the mistakes. More introduction to related works is in Appendix \ref{sec:related}.

\section{The Proposed Framework: {\mymodel}}
As shown in Fig. \ref{fig:framework}, we propose the Divide-Verify-Refine ({\mymodel}) framework which consists of three modules: (a) {\em Divide} instructions and prepare tools accordingly,
(b) {\em Verify} responses and provide feedback,
and (c) {\em Refine} and store responses in a repository.
First, The tool preparation module aims to identify constraints, select appropriate tools, and fill out parameters. In this module, LLMs first decompose the complex instructions into single constraints. For each single constraint, the LLMs will prepare appropriate tools for verification. Second, the prepared tools will verify the response and give detailed textual guidance if the response does not adhere to the constraint. Third, in the self-refinement module, given the textual guidance, LLMs will refine the response to adhere to the target constraint. Since similar few-shot examples usually yield better results, DVR retrieves past refinement experience with the same constraint type few-shot examples. The successfully refined response will be stored for future use. Next, we introduce each module in detail.

\subsection{Divide: Tool Preparation}


To provide accurate feedback, we propose to adopt tools for verification. These tools are widely available and easily accessible \citep{qintoolllm}: (i) There are abundant publicly available tools online. For instance, over 16,000 tools are accessible through RESTful API collections. Additionally, libraries like Regular Expressions (re) and the Natural Language Toolkit (NLTK) \citep{bird2009natural} are commonly used for checking text format, patterns, and length constraints. Hugging Face also provides numerous open-source models, such as topic classifiers \citep{antypas2022twitter} and sentiment classifiers \citep{loureiro2022timelms}, which can be directly integrated and utilized by LLMs; and (ii) When no suitable tool is available, existing works have shown that LLMs can be effectively used to generate reliable tools through code synthesis \citep{guo2024deepseek}. Since tool generation is a one-time cost, it can be efficiently handled by advanced models. In our setup, we use GPT-4o to generate tools, and as shown in Section \ref{sec:toolgen}, our experiments confirm that these generated tools are highly reliable.   

Given an input instruction $I$, the LLM $\mathcal{M}$ first decomposes it into a series of individual constraints. We use a decomposition prompt $p_{decomp}$ asking LLMs for decomposition. With input instruction and decomposition prompt, LLM then generates a set of decomposed constraints: $\mathcal{M}(p_{decomp}, I) \rightarrow \{c_i\}_{i=1,2,3...}$, where $c_i$ is the $i$-th single constraint. For each constraint $c_k$, the LLM determines the appropriate tool by matching $c_k$ to a tool $t_k$ from the predefined toolset: $\mathcal{M}(p_{select}, c_k) \rightarrow t_{k}$, where $t_{k} \in \{t_i\}_{i=1,2,3...}$ is the selected tool for the constraint $c_k$. The prompts for decomposition $p_{decomp}$ and tool selection $p_{select}$ are in Fig. \ref{fig:prompts} in Appendix. After selecting the tools, the LLM sets the necessary parameters for each tool, such as specifying the required number of bullet points or the desired sentiment for the response. Finally, all tools relevant to instruction $I$ are compiled into the set $T_I = \{t_i\}_{i=1,2,3...}$, ready to be utilized in the subsequent verification and feedback phase. 

\subsection{Verify: Verification and Textual Guidance}

Given the instruction, the LLM will first generate the initial response $R_0 = \mathcal{M}(p_{generate}, I)$, where $p_{generate}$ is the prompt for generation (detailed in Fig. \ref{fig:prompts} in Appendix). We denote the current response as $R$ and $R=R_0$ for the first round of refinement and will be updated to the refined response in subsequent rounds. The current response is verified by each tool in toolset $T_I$ as follows:
\begin{equation}
     f_i = t_i(R), \forall t_i \in T_I
\end{equation}
where $f_i$ is the feedback from tool $t_i$ for constraint $c_i$. If the response adheres to the constraint, the feedback is a boolean value ``true''. Otherwise, $f_i$ is a textual feedback that first identifies the error in the response and then suggests modification. For example, as shown in Fig. \ref{fig:framework}, the tool ``Bullet\_points(4)'' counts the number of bullet points in the response and outputs ``true'' if there are 4 bullets; while the response only contains 2 bullets. It finds that the response does not satisfy the constraint and gives out the feedback ``The response only contains 2 bullet points. 2 more bullet points should be added.'' This detailed feedback points out the errors in the response and gives directional information for LLMs to modify the response. We collect all feedback $F_I=\{f_i\}_{i=1,2,3...}$ which will be used to refine the response $R$. 

\subsection{Self-refine with Dynamic Few-shot Prompting}

In the self-refinement phase, the LLM leverages the textual feedback to refine the response. As constraints vary widely, each type of constraint requires demonstrations with similar constraint types for effective refinement. Manually creating one fixed set of few-shot examples can be sub-optimal. Instead of using a fixed set of few-shot examples, we propose dynamic few-shot prompting where few-shot examples with the same constraint type as the current refinement task are selected from the refinement repository. If the response is successfully refined, this process will be stored in the refinement repository for future use.

Specifically, the refinement process targets one unsatisfied constraint at a time, cycling through a refine-verify-refine loop until all constraints are satisfied.  For a given response $R$ and the feedback $f \in F_I, f \neq True$, the refinement response can be written as follows:
\begin{equation}
    R^{\prime} = \mathcal{M}(p_{refine}, s^{t}, I, R, f)
\end{equation}
where $p_{refine}$ is the prompt for refinement (detailed in Fig. \ref{fig:prompts}), $s^{t} = \{(I_i, R_i, f_i, R_i^{\prime})^{t}\}_{i=1,2,3...}$ is the set of refinement examples selected from the refinement repository $Q$, which contains refinement examples having the same constraint type associated with $f$. There might be many refinement examples having the same constraint type with $f$ available in the refinement repository. Retrieval techniques like semantic similarity can be employed to select the most relevant examples. In this paper, we randomly select relevant examples for simplicity and leave more advanced techniques for future work. Some refinement examples are in Fig. \ref{fig:example} in the Appendix.  

If the refined response adheres to the constraint, i.e., $t(R^{\prime}) = True$, the current successful refinement process will be stored in the repository as $Q = Q \cup \{(I, R, f, R^{\prime})^{t}\}$.

\noindent\textbf{Discussion.} Our proposed \mymodel~is a novel approach to enhancing LLMs' ability to follow complex instructions with multiple constraints. The detailed algorithm of \mymodel~is shown in Algorithm \ref{alg: algorithm} in Appendix \ref{sec:algorithm_details}. By integrating external tools for reliable and detailed textual guidance and a refinement repository for storing successful refinement examples, we provide a scalable and robust framework for improving instruction compliance without the need for extensive retraining. Moreover, the external tools and the refinement repository work jointly. Without reliable feedback, the refinement repository would risk accumulating incorrect or noisy examples, which could deteriorate the performance of LLMs over time. The detailed feedback gives ``directional'' information, which guides the LLMs to adjust their responses.  Compared to directly following complex instructions, decomposing these instructions and selecting the appropriate tools are simpler tasks for LLMs. This inherent advantage allows our DVR to be very effective, as it leverages these easier tasks to build a robust system that enhances LLMs' adherence to constraints.

\section{Empirical Validation}
In this section, we conduct experiments to answer the following research questions: (\textbf{RQ1}) Can our {\mymodel} improve the ability of LLMs to follow complex constraints? (\textbf{RQ2}) How does the performance of LLMs differ across various types of constraints, and which constraints pose the greatest challenges? (\textbf{RQ3})  How does each module of {\mymodel} (the tool-assisted verification and the few-shot self-refinement library) individually contribute to improving LLMs' ability to follow constraints? 

\subsection{Experimental Setup}
\textbf{Datasets.} We conduct experiments on two datasets: (i) \textbf{CoDI} \citep{chen2024benchmarking}: A dataset of 500 instructions, each with a topic constraint and a sentiment constraint. (ii) \textbf{ComplexInstruct}: Due to CoDI's limited complexity, we construct ComplexInstruct, a new dataset of complex instructions. Using CoDI's topic instruction set as seed instructions, we refine them by removing implicit length constraints (e.g., replacing "paragraph" or "sentence" with "text") to avoid conflicts and hidden constraints. Then, we synthesize complex instructions by adding constraints to these seed instructions \citep{zhou2023instruction}. To generate instructions of different levels, we generate 6,000 complex instructions across six levels (1–6 constraints per instruction, 1,000 instructions per level). The dataset includes 21 constraint types across 8 general categories (e.g., length, punctuation, case changes), with each type expressed in 8 different ways. The detailed information is in Appendix \ref{sec:ComplexInstruct_info}.

\textbf{Baselines.} We compare our method with state-of-the-art baselines, which can be categorized into three main types: (i) Self-reflection based methods, which iteratively improve response via feedback from LLMs reflection, such as \textbf{Reflexion} \citep{shinn2024reflexion}; (ii) Prompting based methods, which use different prompting strategies to get the best response, including Branch-solve-Merge (\textbf{BSM}) \citep{saha2024branch} and Universal Self-Consistency (\textbf{U-SC}) \citep{chen2024universal}; and (iii) Tool based methods, which use external tools for feedback or selection, such as Rejection sampling (\textbf{R-Sample}) \citep{saunders2022self}, \textbf{React} \citep{yaoreact}, and \textbf{CRITIC} \citep{goucritic}. The details of these baselines are in Appendix \ref{sec:baseline}.


\textbf{Implementation.} We test on popular open-source models including Mistral-7B, Llama3-8B, Llama3.1-8B and Llama3.1-70B. The temperature of the model is 0.8. We set the number of few-shot demonstrations for initial response generation and self-refinement (without repository) as 5 for our method and every baseline. We use the same set of few-shot demonstrations both for baselines and our method. We also set the maximum number of few-shot demonstrations for refinement (with repository) as 8. We set the number of trials as 5 for our method and every baseline. For the refinement repository of our DVR, we consider two variants, i.e., warm-start and cold-start. For \textbf{warm-start}, we have an additional set of instructions (6000 samples for ComplexInstruct and 500 samples for CoDI). Note that these data samples are totally independent with test set. Our framework will first run on these samples to collect examples to fill the refinement repository. For \textbf{cold-start}, since the refinement repository is empty at beginning, we use 5 fixed few-shot examples if there are no examples that can be retrieved from the repository.

\textbf{Evaluation Metrics.} We assess the constraint-following ability by calculating the Instruction Satisfaction Rate (ISR) \citep{jiang2023followbench}. Specifically, each single instruction is satisfied when all constraints in that instruction are satisfied. It is calcualted as $\text{ISR} = \frac{1}{N} \sum_{i=1}^{N} \prod_{j=1}^{m_i} c_{ij}$, where $N$ is the total number of instructions in the dataset, $m_i$ is the number of constraints in the $i$-th instruction, $c_{ij} = 1$ if the $j$-th constraint in $i$-th instruction is satisfied; otherwise $c_{ij}=0$.

\begin{table}[t]
\centering
\caption{Instruction Satisfaction Rate (ISR) across levels 1 to 6 (Llama-3.1-8B-Instruct). The values in parentheses ($+xx$) indicate the improvement compared to the best performing baseline.}   
\label{tab:complex_1}
\vspace*{-1em}
\begin{center}
\resizebox{\columnwidth}{!}{
\begin{tabular}{l c c c c c c }
\hline
Method & Level 1 & Level 2 & Level 3 & Level 4 & Level 5 & Level 6 \\
\hline
Vanilla & 90.5 & 76.6 & 62.5 & 50.1 & 35.6 & 25.3 \\
Reflxion & 91.6 & 78.1 & 63.7 & 49.8 & 35.8 & 25.7 \\
BSM  & 90.1 & 75.3 & 62.0 & 47.5 & 35.5 & 24.1 \\
U-SC & 90.9 & 76.3 & 62.4 & 47.1 & 36.0 & 25.8 \\
R-Sample & 92.1 & 86.7 & 71.1 & 60.4 & 49.8 & 36.3 \\
ReAct & 94.2 & 86.1 & 72.5 & 60.7 & 50.2 & 37.2 \\
CRITIC   & 93.8 & 87.1 & 75.4 & 64.4 & 52.4 & 43.2 \\ %
\hdashline
{\mymodel}$_{CS}$ & 94.5 \scriptsize{(+0.7)} & 87.9 \scriptsize{(+0.8)} & 78.4 \scriptsize{(+3.0)} & 69.5 \scriptsize{(+5.1)} & \textbf{60.9} \scriptsize{(+8.5)} & 49.2 \scriptsize{(+6.0)} \\
{\mymodel}$_{WS}$ & \textbf{95.2} \scriptsize{(+1.4)} & \textbf{88.7} \scriptsize{(+1.6)} & \textbf{79.2} \scriptsize{(+3.8)} & \textbf{69.7} \scriptsize{(+5.3)} & 60.5 \scriptsize{(+8.1)} & \textbf{49.6} \scriptsize{(+6.4)} \\
\hline
\end{tabular}}
\end{center}
\vspace{-1mm}
\end{table}

\begin{table}[t]

\centering
\caption{Instruction Satisfaction Rate (ISR) across levels 1 to 6 
 (Mistral-7B-Instruct-v0.3)}
\label{tab:complex_2}
\vspace*{-1em}
\begin{center}
\resizebox{\columnwidth}{!}{
\begin{tabular}{ l c c c c c c }
\hline
Method & Level 1 & Level 2 & Level 3 & Level 4 & Level 5 & Level 6 \\
\hline
Vanilla & 77.0 & 55.3 & 34.1 & 19.9 & 12.4 & 6.3 \\
Reflxion & 77.2 & 55.8 & 35.1 & 20.1 & 12.0 & 5.8 \\
BSM & 78.1 & 56.2 & 33.8 & 19.3 & 11.3 & 5.2 \\
U-SC  & 76.8 & 56.0 & 34.3 & 20.4 & 12.9 & 5.8 \\
R-Sample & 78.4 & 58.3 & 37.6 & 23.0 & 13.5 & 6.8 \\
ReAct   & 86.0 & 67.8 & 46.0 & 32.5 & 18.2 & 10.7 \\
CRITIC  & 88.9 & 72.5 & 55.6 & 43.5 & 28.1 & 18.1 \\
\hdashline
{\mymodel}$_{CS}$ & 94.9 \scriptsize{(+6.0)} & 80.2 \scriptsize{(+7.7)} & 64.1 \scriptsize{(+8.5)} & 49.3 \scriptsize{(+5.8)} & 35.8 \scriptsize{(+7.7)} & \textbf{23.6} \scriptsize{(+5.5)}  \\
 {\mymodel}$_{WS}$ & \textbf{95.0} \scriptsize{(+6.1)} & \textbf{81.3} \scriptsize{(+8.8)} & \textbf{66.6} \scriptsize{(+11.0)} & \textbf{51.4} \scriptsize{(+7.9)} & \textbf{36.4} \scriptsize{(+8.3)} & 23.4 \scriptsize{(+5.3)} \\
\hline
\end{tabular}}
\end{center}
\end{table}

\begin{figure*}[t]
\centering

\includegraphics[width=0.24\textwidth]{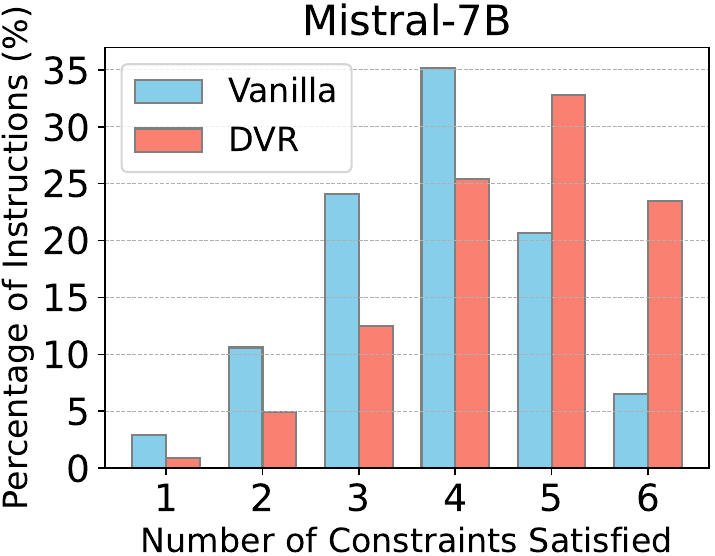} \hfill
\includegraphics[width=0.24\textwidth]{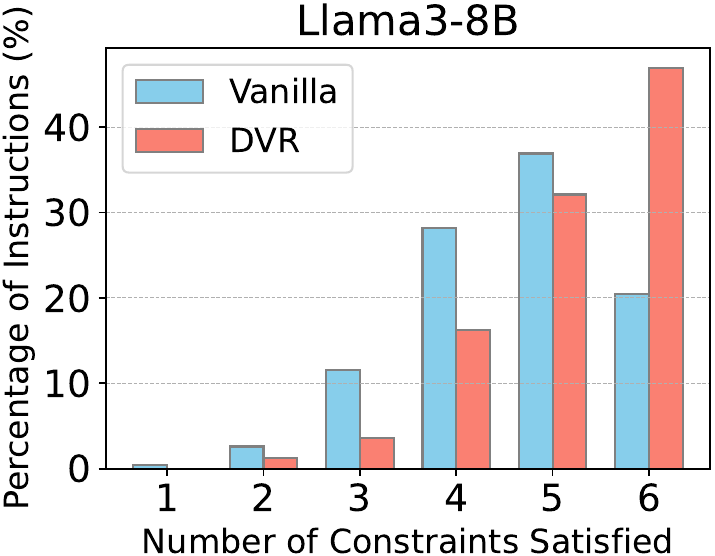} \hfill
\includegraphics[width=0.24\textwidth]{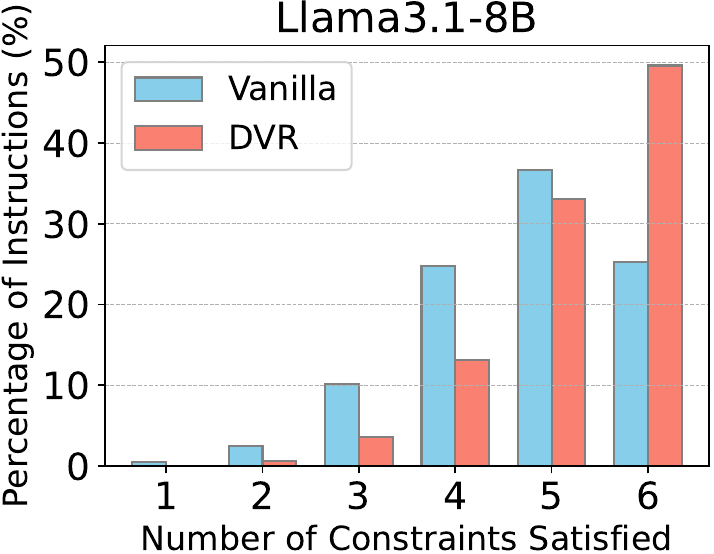} \hfill
\includegraphics[width=0.24\textwidth]{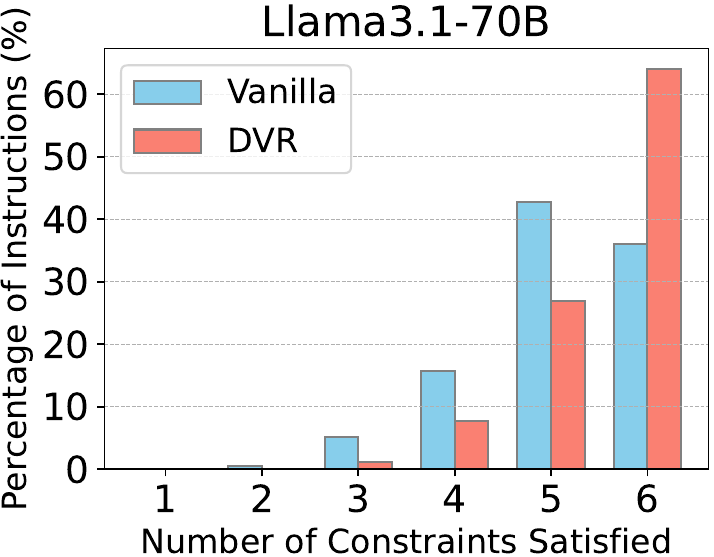}
\vskip -0.5em
\caption{Distribution of satisfied constraints number per instruction (level 6).}
\label{fig:distribution}
\end{figure*}

\begin{table}[t]
\centering
\caption{Performance by Self-training (zero-shot)}
\label{tab:DPO}
\vspace*{-1em}
\centering
\resizebox{\columnwidth}{!}{\begin{tabular}{ l c c c c c c }
\hline
Model & Level 1 & Level 2 & Level 3 & Level 4 & Level 5 & Level 6 \\
\hline
Mistral-7B & 60.1 & 41.7 & 23.8 & 14.2 & 8.4 & 3.7 \\
Mistral-7B(DPO) & 85.0 & 67.8 & 47.0 & 33.9 & 18.4 & 12.9 \\
Llama3.1-8B & 62.5 & 38.9 & 23.2 & 14.6 & 9.6 & 6.3 \\
Llama3.1-8B(DPO) & 83.7 & 69.8 & 55.6 & 39.3 & 28.0 & 17.0 \\

\hline
\end{tabular}}
 
\end{table}


\subsection{RQ1: Constraint-Following Ability}
To answer RQ1, we evaluate DVR on two datasets. We evaluate structural constraints (e.g., text length, number of sections, and bullet points) on ComplexInstruct and content constraints (e.g., topic and sentiment constraints) on CoDI respectively. {\mymodel}$_{CS}$ and {\mymodel}$_{WS}$ are cold-start and warm-start. 

Results are shown in Table \ref{tab:complex_1} and Table \ref{tab:complex_2} \textbf{(i) Single vs Multi-constraints:} As constraint complexity increases, satisfaction rates drop. While ISR at Level 1 approaches 95\%, Level 6 ISR drops to 25\% for Llama3.1-8B and 6.3\% for Mistral-7B, showing LLMs struggle with multiple constraints even when they handle them individually. \textbf{(ii) Intrinsic self-reflection is unreliable:} Reflxion \citep{shinn2024reflexion}, which relies on LLMs to reflect and self-correct, shows minimal improvement over Vanilla, indicating LLMs struggle to identify their own constraint violations. Similar results can be observed on Branch-Solve-Merge and Universal Self-consistence. \textbf{(iii) Textual Guidance matters:} ReAct \cite{yaoreact} and CRITIC \cite{goucritic} can be viewed as two variants of DVR, where feedback is provided as boolean signals on the instruction level or specific constraint violations. Compared with ReAct and CRITIC, DVR has better performance, which means detailed analysis and textual guidance can make the refinement more effective. The refinement repository further maximizes the refinement effectiveness. The results in Figure \ref{fig:distribution} show the distribution of satisfied constraints per instruction at Level 6 difficulty. Our framework shifts the distribution rightward, indicating improved adherence to multiple constraints. Notably, for Mistral-7B, our framework moves the central tendency from satisfying 4 constraints to 5.

\textbf{Self-improving:} To enable self-improvement without relying on labeled data or external models, we leverage the output of \mymodel~ as training data. The refined response and the original model response are selected as positive-negative a pair if their constraint satisfaction rate gap is over 0.4. These pairs are further used for Direct Preference Optimization (DPO) tuning \citep{rafailov2024direct}. This allows the model to iteratively enhance its performance based on its outputs. For fine-tuning, we use Llama3.1-8B and Mistral-7B as the base models and apply LoRA \citep{hu2021lora} with a rank of 32 and an $\alpha$ value of 64. For DPO tuning, $ \beta $ is set as 0.2.

The results, presented in Table \ref{tab:DPO}, demonstrate that DPO-tuned models (DPO-DVR) significantly outperform the vanilla model across all constraint levels, particularly at higher complexity levels. This highlights the effectiveness of the self-training approach in improving constraint adherence. 

\textbf{Additional Results}: Additional experimental results on models such as LLaMA 3.1-70B, LLaMA 3-8B, and GPT-4 can be found in Appendix \ref{sec:detaled_experiment}, along with some evaluations on the CoDI \citep{chen2024benchmarking} and IFeval \citep{zhou2023instruction} benchmarks. We also test DVR efficiency by testing the time used for inference and results are in Appendix \ref{sec:time}. Discussion on the influence of DVR on fluency and readability can be found in \ref{sec:descriptive}.

\begin{table}[t]
\centering
\caption{Comparison Across Constraints Types}
\label{tab:type_comparison}
\vspace*{-1em}
\begin{center}
\resizebox{\columnwidth}{!}{
\begin{tabular}{l cc cc cc}
\toprule
& \multicolumn{2}{c}{Mistral-7B} & \multicolumn{2}{c}{Llama3-8B} & \multicolumn{2}{c}{Llama3.1-8B}  \\
\cmidrule(lr){2-3} \cmidrule(lr){4-5} \cmidrule(lr){6-7} 
Constraint Type & Vanilla & \mymodel & Vanilla & \mymodel & Vanilla & \mymodel  \\
\midrule
Detectable Content & 76.36 & 88.90 & 84.18 & 96.81 & 86.29 & 96.31  \\
Keywords            & 76.04 & 84.23 & 83.84 & 88.32 & 84.94 & 88.77  \\
Punctuation         & 24.34 & 72.93 & 91.03 & 95.64 & 97.01 & 98.04  \\
Case Change        & 70.08 & 81.28 & 81.28 & 93.38 & 82.97 & 90.71  \\
Start End            & 81.29 & 90.41 & 84.88 & 90.03 & 84.07 & 91.92  \\
Detectable Format  & 69.59 & 80.70 & 81.57 & 89.23 & 84.69 & 92.31 \\
Language            & 69.11 & 81.80 & 77.06 & 88.38 & 81.96 & 89.76  \\
Length Constraints & 50.42 & 73.23 & 65.29 & 80.85 & 68.55 & 83.57  \\
\bottomrule
\end{tabular}}
\end{center}
\vskip -1em
\end{table}

\subsection{RQ2: Comparison Across Different Constraint Types}
Comparison across different constraint types is shown in Table \ref{tab:type_comparison} (warmstart). Coldstart results are provided in Table \ref{tab:type_comparison_cold} in Appendix \ref{sec:detaled_experiment}. We have the following observations. (i) Length constraints are the most challenging: Every model struggles with length constraints, which include minimum/maximum word counts, sentence counts, and exact paragraph requirements. This difficulty likely stems from a lack of such constraints in instruction-tuning datasets, making it hard for models to map output structure to length requirements. Additionally, length constraints require models to plan responses from the outset while maintaining coherence and completeness. (ii) Language constraints, which require the use of languages such as Italian, German, or Japanese, are the second most challenging. This might be due to the limited multilingual capabilities of the LLMs. (iii) Punctuation constraint which requires LLMs not to use any commas in their responses, is especially challenging for Mistral-7B. However, our framework improves it significantly and triples the performance from 24\% to 73\%. DVR's feedback not only verifies correctness but also explicitly highlights the locations of commas, providing precise guidance.

\subsection{RQ3: Contribution of Individual Modules}

\begin{figure}[h]
\centering
\includegraphics[width= \columnwidth]{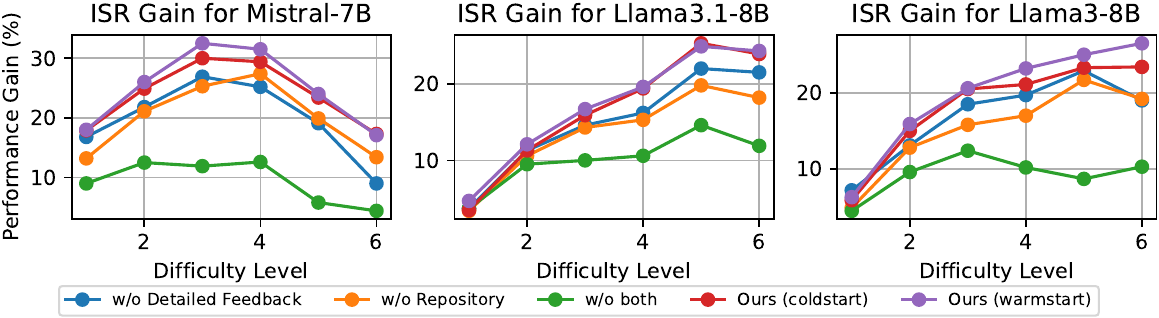}  
\caption{Ablation study on Mistral-7B, Llama3.1-8B and Llama3-8B.}
\label{fig:Ablation}
\end{figure}

We conduct an ablation study to evaluate the impact of key components. We examine three variants: (i) \textbf{w/o Detailed Feedback:}Removes detailed feedback from the tool but retains the refinement repository, which provides relevant few-shot examples showing responses before and after refinement. The repository starts empty (coldstart).  (i) \textbf{w/o Repository:} Removes the refinement repository, using only five fixed examples for self-refinement. (i) \textbf{w/o both:} The refinement repository and detailed feedback are all removed. Figure \ref{fig:Ablation} shows performance gaps between each method and the Vanilla. Both detailed feedback and the refinement repository are crucial. Without the repository, performance gains are limited, as fixed few-shot examples are suboptimal for diverse refinement needs. Detailed feedback is essential as it pinpoints errors and provides direction for response modification. 

\subsection{Hyper-Parameter Sensitivity Analysis}

\begin{figure}[t]
\centering
\includegraphics[width=0.43\columnwidth]{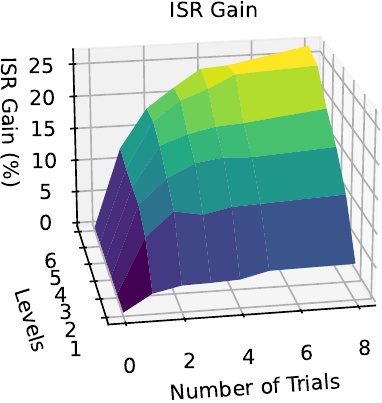} \hfill
\includegraphics[width=0.48\columnwidth]{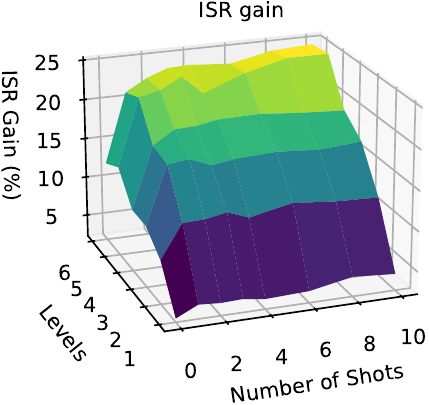} 

\caption{Parameter study on ComplexInstruct.} 
\label{fig:parameter_study}

\end{figure}

We also conduct a hyper-parameter sensitivity analysis of our framework, testing different numbers of refinement few-shots and trials for successful refinement on Llama3.1-8B. As shown in Figure \ref{fig:parameter_study}, performance improves with more trials but saturates at five, with minimal gains beyond that. Similarly, increasing the few-shot examples boosts performance in the beginning. The performance saturates after 8 shots. This indicates that the first few numbers of trials and few-shot examples are most effective for refining the response.

\begin{table}[h]
 
\centering
\caption{LLMs Performance on Tool Selection. (HL: Hamming Loss)}
\vspace*{-1em}
\begin{center}
\resizebox{\columnwidth}{!}{
\begin{tabular}{lrrrrr}
\toprule
      Models &  HL &  Acc &  Precision &  Recall &  F1 \\
\midrule
     Mistral-7B &          4.13 &     52.85 &      92.98 &   81.39 &     86.80 \\
   Llama3-8B &          2.64 &     67.60 &      94.39 &   89.48 &     91.87 \\
 Llama3.1-8B &          2.90 &     61.77 &      94.69 &   87.50 &     90.95 \\
Llama3.1-70B &          0.86 &     86.40 &      98.41 &   96.38 &     97.38 \\
\bottomrule
\end{tabular}}
\label{tab:tool_selection}

\end{center}
 \vspace*{-1em}
\end{table}

\subsection{Tool Selection Accuracy}
Correctly decomposing and selecting tools are essential for feedback and refinement. We define tool selection as a multi-label prediction task for LLMs, evaluated using hamming score, accuracy, precision, recall, and F1-score. The total number of tools is 21. Results are shown in Table \ref{tab:tool_selection}. Hamming loss, which measures the fraction of incorrect labels, is low across all models, indicating minimal mispredictions. Every model demonstrates a very high precision score, meaning that the tools they select are mostly correct, avoiding misleading feedback with incorrect tool selection. Accuracy, which measures the exact match between the selected tools and the ground truth, is the strictest metric. Despite this, all models achieve over 50\% accuracy. Considering the limited performance of these models on constraint-following tasks, tool selection is a relatively easier task for LLMs. This performance gap makes it possible for our method to provide reliable feedback, collect past refinement examples and be effective in improving LLMs' constraint-following ability. We also evaluate DVR's robustness to tool errors, with detailed experiments in Appendix \ref{sec:robustness}.

\subsection{LLM Self-verify Ability}
We evaluate the LLM ability to verify whether the responses meet the given constraints. Specifically, we present response-constraint pairs and ask LLMs to determine if the response aligns with the constraint. As shown in Table \ref{tab:self_verify_acc}, the verification accuracy is around 0.5, similar to random guessing. This suggests that LLMs struggle to accurately assess responses, making them unreliable for self-feedback. In contrast, compared with LLM self-verification ability, LLMs perform significantly better in tool selection. This performance gap ensures the effectiveness of DVR. Future work can explore this further, including developing benchmarks for scenarios involving thousands of tools.
 
\begin{table}[t]
 
\centering
\caption{LLMs Self-verification Accuracy (\%)}
 \vspace*{-1em}
\label{tab:self_verify_acc}
\begin{center}
\resizebox{\columnwidth}{!}{
\begin{tabular}{ l c c c}
\hline
Model &  Mistral-7B & Llama3-8B & Llama3.1-8B  \\
& 53.1 & 56.8 & 55.7 \\

\hline
\end{tabular}}
\end{center}
 
\end{table}

\subsection{Tool Generation Accuracy}
\label{sec:toolgen}
We use GPT-4o to generate Python scripts as tools, testing all 21 types of tools. Among them, 20 functioned correctly. One script designed to check the existence of a title fails for an edge case where the title is blank ("<<>>"). This demonstrates that tool generation is overall reliable and scalable. To balance reliability and cost, a practical strategy is to generate tools with GPT-4o and save them locally for reuse, reducing API costs. Some tool examples are shown in Appendix Fig \ref{fig:tools}.

\section{Conclusion}
We propose the Divide-Verify-Refine (DVR) framework to enhance LLMs' ability to follow multi-constraint instructions. DVR has three steps: (1) Divide complex instructions into single constraints and assign appropriate tools for each constraint. (2) Verify: To tackle the feedback quality problem, these tools rigorously verify the response and generate textual guidance for refinement. (3) Refine: To maximize the refinement effectiveness, we design the dynamic few-shot prompting with a refinement repository to store past refinement experiences. DVR improves LLMs' adherence to complex multi-constraint instructions. Additionally, we construct a new dataset free from hidden or conflicting constraints, providing a more comprehensive and accurate evaluation of LLM performance on multi-constraint following.

\section{Limitations}

There are several limitations and potential future works. 
(1) Currently, we consider multiple independent constraints. However, the instructions in real-world might be more complex and constraints might have dependency with each other \citep{wen2024benchmarking}. For example, the instruction can ask the response to have 4 bullet points and 2 sentences in each bullet point. In such a scenario, LLMs need to assign different priorites of these constraints. (2) Moreover, tools may not be available for new constraints. Here, we assume that we have tools for all existing constraints. However, users' requirements can be very diverse and we would not have certain tools for new constraints. (3) As shown in Appendix \ref{sec:robustness}, DVR performance would decline when tools produce errors.

\bibliography{acl_latex}

\appendix

\section{Appendix}

\subsection{Algorithm}
\label{sec:algorithm_details}
DVR algorithm is shown in Algorithm \ref{alg: algorithm}. The process begins with the LLM selecting the appropriate tools for each instruction. The selected tools assess whether the generated response meets the constraints and provide textual feedback if any violation is detected. To improve adherence, few-shot examples with the same constraint type are retrieved from a refinement repository. If the response is successfully refined and passes the tool-based validation, this refined version is stored in the repository for future use. The response is output when it passes all tools or the budget is met.

\begin{algorithm}[h]
\caption{Algorithm for \mymodel}
\label{alg: algorithm}
\textbf{Input:} LLM $\mathcal{M}$, Instructions $X$, Toolset $T$ \\
\textbf{Output:} Response set $Y$ \\
\textbf{Select:} Trial number $n$
\begin{algorithmic}[1]

\STATE Initialize the refinement repository $Q = \{\}$
\FOR{$I \in X$}
\STATE Generate response: $R_0 = \mathcal{M}(p_{generate}, I)$.
\STATE Initialize the toolset for $I$: $T_I = \{\}$.
\STATE Decompose:$\mathcal{M}(p_{decomp}, I) \rightarrow \{c_i\}_{i=1,2...}$
\FOR{$c \in \{c_i\}_{i=1,2,3...}$}
\STATE $\mathcal{M}(p_{select}, c) \rightarrow t$, where $t \in T$.
\STATE $\mathcal{M}$ sets parameters for $t$.
\STATE $ T_I = T_I  \cup t $.
\ENDFOR
\STATE $R=R_0$, $a=n$.
\WHILE{$a > 0$}
\STATE $a=a-1$
\STATE Verify and get feedback from tools: $F_I=\{f_i\}_{i=1,2,3...}$, where $f_i = t(R)$.

\IF{$f=True, \forall f \in F_I$}
\RETURN $R$
\ENDIF

\STATE Retrieve: $s^{t} = \{(I_i, R_i, f_i, R_i^{\prime})^{t}\}_{i=1,2...}$, where $s^{t} \subseteq  Q$.
\STATE Refine: $R^{\prime} = \mathcal{M}(p_{refine}, s^{t}, I, R, f)$, where $f \in F_i$ and $f \neq True$.
\IF{$t(R^{\prime})=True$}
\STATE Save: $Q=Q \cup \{(I, R, f, R^{\prime})^{t}\}$
\STATE Update the response: $R=R^{\prime}$
\STATE $a=n$
\ENDIF

\ENDWHILE
\STATE $Y = Y \cup R $
\ENDFOR
\end{algorithmic}
\end{algorithm}


\subsection{Baseline Details}
\label{sec:baseline}
We conduct experiments on 6 baselines, encompassing reflexion-based approaches, prompting strategies, and tool-assisted techniques. The details are as follows:

\begin{itemize}
\item Reflexion \citep{shinn2024reflexion}: This method allows LLMs to self-reflect on their own responses and provide valuable feedback for future outputs. With the feedback, LLMs will refine their responses. 
\item Branch-solve-Merge (BSM) \citep{saha2024branch}: BSM uses a "Divide and Conquer" approach to break complex instructions as individual branches. Then the LLMs will merge the responses from branches as the final answer. Similarly, in our experiment, we use LLMs to generate a response for each single constraint and then merge them together.

\item Universal Self-Consistency (U-SC) \citep{chen2024universal}: This study extends the idea of Self-Consistency \citep{wangself} to free-form generation. It first generates several candidate responses and then asks LLMs to select the most consistent one.

\item Rejection Sampling \citep{saunders2022self}: Since we have tools for reliable verification, the most simple method is to select the best one from a set of responses. Here, we give the maximum number of trials as 5.

\item ReAct \citep{yaoreact}: In ReAct, LLMs take actions based on the observation of the environment. Here, we adopt this method by letting the tools as the environment and giving LLMs boolean signals indicating whether the generated response adheres to all constraints in the instruction.

\item CRITIC \citep{goucritic}: CRITIC uses external API to evaluate the toxicity score of a generated response, focusing on a single predefined task. We adopt this method as a variant of our DVR framework, where tools will pinpoint which constraint of the instruction is not satisfied.

\end{itemize}

\subsection{ComplexInstruct}
\label{sec:ComplexInstruct_info}

We have 21 types of constraints which can be divided into 8 general categories \citep{zhou2023instruction} as shown below:

\begin{itemize}
    \item Keywords: 
    
    (1) Include keyword, \\(2) Include keyword at least/less than certain frequency,\\ (3) Forbidden word,\\ (4) At least/less than certain frequency of letters. 
    
    \item Length: 
    
    (1) At least/less than certain number of words, \\(2) At least/less than certain number of sentences, \\(3) Exact number of paragraphs.

    \item Detectable Content: \\(1) postscript, \\(2) Exact number of placeholders.

    \item Detectable Format: \\(1) Number of bullet points, \\(2) Add title, \\(3) Answer from options, \\(4) Minimum of highlighted sections,\\ (5) Json format.

    \item  Change Cases: \\(1) All uppercase, \\(2) All lowercase, \\(3) At least/less than certain number of all-capital words.

    \item Startend: \\(1) End the text with a certain sentence, \\(2) Wrap whole response in double quotation.

    \item  Punctuation: \\(1) No commas in response.

    \item  Language: \\(1) Respond with certain language.
\end{itemize}

\subsection{Detailed experiments}
\label{sec:detaled_experiment}

Additional experiments on Llama3-8B and Llama3.1-70B are shown in Table \ref{tab:Llama-3-8B} and Table \ref{tab:Llama-3.1-70B} respectively. We can observe that our methods consistently outperform baselines on different LLMs. In Table \ref{tab:topic_sentiment}, we also observe that LLMs perform overall good on the CoDI dataset \cite{chen2024benchmarking}. There are two reasons. The first reason is that instructions are relatively simple, and only contain two constraints. Additionally, another study also shows that LLMs perform relatively better on sentiment and topic constraints \cite{chen2024benchmarking} compared with format constraints. The LLMs inherently have better performance on semantic constraints over structural constraints. Our methods also outperform baselines and successfully improve the instruction satisfaction rate on CoDI.


\begin{table}[h]
\centering
\caption{Performance of methods across levels 1 to 6 (Llama-3-8B-Instruct)}
\begin{center}
\resizebox{\columnwidth}{!}{
\begin{tabular}{ l c c c c c c }
\hline
Method & Level 1 & Level 2 & Level 3 & Level 4 & Level 5 & Level 6 \\
\hline
Vanilla & 89.1 & 71.7 & 56.4 & 44.2 & 30.4 & 20.4 \\
Reflxion  & 88.8 & 72.1 & 57.5 & 41.5 & 30.0 & 20.9 \\
BSM & 89.2 & 71.9 & 56.0 & 41.3 & 28.8 & 19.5 \\
U-SC   & 89.5 & 71.8 & 56.7 & 45.1 & 31.2 & 20.6 \\
R-Sample & 90.8 & 80.9 & 64.5 & 52.9 & 39.8 & 31.0 \\
ReAct & 93.6 & 81.3 & 68.8 & 54.4 & 39.1 & 30.7 \\
CRITIC  & 94.1 & 85.8 & 74.4 & 61.2 & 51.1 & 41.5 \\ \hdashline
\mymodel$_{CS}$ & 95.0 & 86.7 & 76.9 & 65.3 & 53.7 & 43.8 \\
\mymodel$_{WS}$ & 95.4 & 87.6 & 77.0 & 67.4 & 55.4 & 46.9 \\
\hline
\end{tabular}}
\label{tab:Llama-3-8B}
\end{center}
\end{table}

\begin{table}[h]
 
\centering
\caption{Performance of methods across levels 1 to 6 (Llama-3.1-70B-Instruct-AWQ-INT4)}

\begin{center}
\resizebox{\columnwidth}{!}{
\begin{tabular}{ l c c c c c c }
\hline
Method & Level 1 & Level 2 & Level 3 & Level 4 & Level 5 & Level 6 \\
\hline
Vanilla & 95.5 & 83.7 & 72.4 & 63.2 & 51.3 & 35.9 \\
Reflexion & 95.3 & 83.5 & 72.8 & 63.0 & 51.6 & 36.1 \\
BSM & 95.0 & 84.5 & 72.6 & 64.8 & 49.6 & 34.2 \\
U-SC & 96.0 & 83.3 & 71.8 & 64.0 & 52.5 & 36.3 \\
R-Sample & 97.3 & 90.8 & 83.2 & 72.2 & 63.8 & 50.7 \\
ReAct & 97.5 & 91.0 & 83.5 & 72.4 & 65.0 & 52.1 \\
CRITIC & 98.1 & 93.2 & 87.6 & 79.1 & 73.7 & 61.3 \\ \hdashline
\mymodel$_{CS}$ & 98.0 & 94.3 & 88.2 & 82.0 & 75.7 & 63.1 \\
\mymodel$_{WS}$ & 98.2 & 94.6 & 88.7 & 82.2 & 76.0 & 64.2 \\
\hline
\end{tabular}}
\end{center}
\label{tab:Llama-3.1-70B}
\end{table}

\begin{table}[t]
\centering
\caption{Instruction Satisfaction Rate (ISR) on CoDI}
\label{tab:topic_sentiment}
\centering
\resizebox{\columnwidth}{!}{
\begin{tabular}{l c c c }
\hline
Method & Mistral 7B & Llama3 8B & Llama3.1 8B \\ \hline
Vanilla      & 68.8 & 68.8 & 68.6 \\ 
Reflexion            & 69.4 & 70.0 & 69.8 \\ 
BSM      & 68.2 & 68.4 & 68.6 \\ 
USC & 69.2 & 70.2 & 69.6 \\ 
Reject Sample      & 79.8 & 80.8 & 81.4 \\ 
ReAct                   & 80.4 & 81.0 & 81.8 \\ 
CRITIC                 & 88.6 & 93.0 & 91.2 \\ 
\hdashline
{\mymodel} (coldstart)                    & 92.0 & 94.2 & \textbf{94.6} \\ 
{\mymodel} (warmstart)                    & \textbf{93.2} & \textbf{94.4} & \textbf{94.6} \\ \hline
\end{tabular}
}

\end{table}

\begin{table}[h]
 
\centering
\caption{Comparison Across Different Constraints Types (coldstart)}
\label{tab:type_comparison_cold}
\begin{center}
\resizebox{\columnwidth}{!}{
\begin{tabular}{l cc cc cc cc}
\toprule
& \multicolumn{2}{c}{Mistral-7B} & \multicolumn{2}{c}{Llama3-8B} & \multicolumn{2}{c}{Llama3.1-8B} & \multicolumn{2}{c}{Llama 3.1-70B} \\
\cmidrule(lr){2-3} \cmidrule(lr){4-5} \cmidrule(lr){6-7} \cmidrule(lr){8-9}
Constraint Type & Vanilla & \mymodel & Vanilla & \mymodel & Vanilla & \mymodel & Vanilla & \mymodel \\
\midrule
Detectable Content & 76.36 & 88.49 & 84.18 & 95.82 & 86.29 & 96.19 & 97.06 & 98.14 \\
Keywords            & 76.04 & 84.32 & 83.84 & 87.53 & 84.94 & 88.77 & 87.88 & 92.05 \\
Punctuation         & 24.34 & 71.31 & 91.03 & 95.47 & 97.01 & 98.29 & 98.38 & 98.38 \\
Case Change        & 70.08 & 80.15 & 81.28 & 93.20 & 82.97 & 89.96 & 80.23 & 96.24 \\
Start End            & 81.29 & 88.71 & 84.88 & 88.71 & 84.07 & 91.78 & 89.37 & 95.94 \\
Detectable Format  & 69.59 & 81.30 & 81.57 & 89.29 & 84.69 & 92.38 & 90.52 & 95.56 \\
Language            & 69.11 & 82.72 & 77.06 & 86.24 & 81.96 & 89.30 & 90.83 & 94.34 \\
Length Constraints & 50.42 & 72.61 & 65.29 & 79.66 & 68.55 & 83.93 & 80.50 & 90.17 \\
\bottomrule
\end{tabular}}
\end{center}

\end{table}

\textbf{Experiments on IFEval:} We conduct experiments on IFEval \citep{zhou2023instruction} which is an instruction-following benchmark widely used for industry. The IFEval dataset evaluates the instruction-following ability and is one of the core benchmarks used in the Open LLM Leaderboard (Hugging Face). We conduct experiments on Mistral-7B-v0.3 and the results are shown in Table \ref{tab:ifeval}. DVR outperforms all other baselines on IFEval benchmark.

\begin{table}[H]
\centering
\caption{ISR (\%) for IFEval Dataset}
\label{tab:ifeval}
\centering
\resizebox{\columnwidth}{!}{\begin{tabular}{ l c c c c c c c c }
\hline
Method & Vanilla & Reflexion & BSM & U-SC & R-Sample & ReAct & CRITIC & DVR \\
\hline
ISR & 47.32 & 47.13 & 47.87 & 46.95 & 53.23 & 53.97 & 55.53 & 60.44 \\
\hline
\end{tabular}} 
\end{table}

\textbf{Experiments on GPT4:} We conduct experiments on GPT-4-turbo. Shown in Table \ref{tab:GPT4}, we can observe that GPT-4-turbo performs better than open-source models (Mistral and Llama). Surprisingly, applied on Llama3.1-8B, DVR can still outperform GPT-4-turbo, indicating that DVR exploits the potential of the open-source model.

\begin{table}[H]
\centering
\caption{Performance Comparison to GPT4-turbo (zero shot)}
\label{tab:GPT4}
\centering
\resizebox{\columnwidth}{!}{\begin{tabular}{ l c c c c c c }
\hline
Model & Level 1 & Level 2 & Level 3 & Level 4 & Level 5 & Level 6 \\
\hline
Mistral-7B & 77.0 & 55.3 & 34.1 & 19.9 & 12.4 & 6.3 \\
DVR (Mistral-7B) & 95.0 & 81.3 & 66.6 & 51.4 & 36.4 & 23.4 \\
Llama3.1-8B & 90.5 & 76.6 & 62.5 & 50.1 & 35.6 & 25.3 \\
DVR (Llama3.1-8B) & 95.2 & 88.7 & 79.2 & 69.7 & 60.5 & 49.6 \\
GPT4-turbo(zero shot) & 95.3 & 88.4 & 78.8 & 65.2 & 53.7 & 42.6 \\
\hline
\end{tabular}}
 
\end{table}

\subsection{Fluency and Readability}
\label{sec:descriptive}

In this subsection, we investigate if our framework would sacrifice comprehensibility and fluency in order to follow complex-constraints. We evaluate key metrics such as readability, perplexity, and coherence. These metrics assess the comprehensibility and fluency of the responses. Results of ComplexInstruct and CoDI are shown in Table \ref{tab: quality_structural} and Table \ref{tab:descriptive_codi}. They both show that our framework has performance comparable to those of Vanilla, indicating that it does not degrade fluency and readability. The reason is that our method does not change any weights in LLMs, which maintains their ability in generating fluent and comprehensible text.

\begin{table}[H]

\centering
\caption{Descriptive Statistics of Responses (ComplexInstruct), where Coherence\_or1 is first order coherence and Coherence\_2 is the second order coherence.}
\begin{center}
\resizebox{\columnwidth}{!}{
\begin{tabular}{l  l c c c c }

\hline
Model & Method &  Readability $\uparrow$ & Perplexity $\downarrow$ & Co\_or1  $\uparrow$  & Co\_or2 $\uparrow$ \\ \hline
\multirow{2}{*}{mistral7B}      & Vanilla             & 62.24                & 18.22              & 0.61                           & 0.59                            \\ 
               & \mymodel            & 61.93                & 18.95              & 0.59                           & 0.57                            \\
\multirow{2}{*}{llama3-8B}     & Vanilla             & 63.77                & 18.49              & 0.57                           & 0.57                            \\
               & \mymodel            & 63.58                & 18.08              & 0.59                           & 0.56                            \\
\multirow{2}{*}{llama3.1-8B}    & Vanilla             & 63.43                & 19.68              & 0.62                           & 0.61                            \\ 
               & \mymodel            & 62.75                & 18.30              & 0.62                           & 0.60                            \\ 
\multirow{2}{*}{llama3.1-70B}   & Vanilla             & 61.96                & 17.99              & 0.64                           & 0.63                            \\ 
               & \mymodel            & 63.51                & 18.04              & 0.63                           & 0.62                            \\ \hline
\end{tabular} }
\end{center}
\label{tab: quality_structural}
\vskip -1em
\end{table}

\begin{table}[H]
\vskip -1em
\centering
\caption{Descriptive Statistics of Responses (CoDI)}
 
\centering
\resizebox{\columnwidth}{!}{
\begin{tabular}{l l c c c c }
\hline
Model & Method &  Readability $\uparrow$ & Perplexity $\downarrow$ & Co\_or1  $\uparrow$  & Co\_or2 $\uparrow$ \\ \hline
\multirow{2}{*}{mistral7B} & Vanilla  & 63.62 & 15.27 & 0.82 & 0.81 \\
                           & \mymodel & 63.53 & 15.98 & 0.83 & 0.82 \\ 
\multirow{2}{*}{llama3-8B} & Vanilla  & 63.79 & 14.25 & 0.79 & 0.76 \\
                           & \mymodel & 64.14 & 16.07 & 0.80 & 0.77 \\
\multirow{2}{*}{llama3.1-8B}& Vanilla  & 62.18 & 16.27 & 0.81 & 0.83 \\ 
                           & \mymodel & 62.02 & 17.58 & 0.81 & 0.80 \\ \hline
\end{tabular}}
\label{tab:descriptive_codi}
 
\end{table}



\subsection{Robustness of DVR}
\label{sec:robustness}
We also conduct experiments to assess DVR’s performance in the presence of tool errors. Two types of errors are introduced: random noise and systematic bias. Specifically, we evaluate the framework on instructions with length constraints, using 600 samples (from ComplexInstruct) for word count control and another 600 for sentence count control. A constraint example: “The response needs to be less than (or at least) x number of words/sentences.” where x ranges from 10 to 100 for words and 3 to 5 for sentences. We add two types of noises to tools: 

\textbf{Noise:} Gaussian noise with a mean of 0 is added to the counted number of words (or sentences) to simulate random errors. The DVR’s performance is then measured across different deviation levels.

\textbf{Bias Errors:} A fixed bias is added to the counted values of words (or sentences) to introduce systematic errors. The tables below demonstrate DVR’s performance under different bias values.

\textbf{Observations:} We have several observations in Table \ref{tab:error_words} and Table \ref{tab:error_sentences}. (1) The performance will decrease as the noise levels (deviation, bias values) increase. (2) As the errors become large, the performance degradation will saturate. (3) Overall, DVR will not perform much worse than vanilla even if the bias and errors are large (20 for word count and 4 for sentence count). (4) The impact of noise on the overall instruction satisfaction rate is less severe compared to its influence on specific constraints.

\begin{table}[H]
\caption{Satisfaction Rate for Word Number Constraints (\%)}
\label{tab:error_words}
    \centering
    \resizebox{\columnwidth}{!}{\begin{tabular}{lccccc}
        \toprule
        Deviation & 0 & 5 & 10 & 20 & Vanilla \\ \midrule
        Words Number Satisfaction Rate & 88.00 & 87.17 & 82.83 & 81.50 & 68.16 \\
        Instruction Satisfaction Rate & 48.17 & 45.00 & 43.67 & 43.67 & 10.17 \\ \bottomrule
    \end{tabular}}
    \vspace{1em}
    
    \resizebox{\columnwidth}{!}{\begin{tabular}{lccccc}
        \toprule
        Bias & 0 & 5 & 10 & 20 & Vanilla \\ \midrule
        Words Number Satisfaction Rate & 88.00 & 87.17 & 84.33 & 82.67 & 68.16 \\
        Instruction Satisfaction Rate & 48.17 & 47.83 & 47.00 & 45.67 & 10.17 \\ \bottomrule
    \end{tabular}}
\end{table}

\vskip -2em
\begin{table}[H]
\caption{Satisfaction Rate for Sentence Number Constraints (\%)}
\label{tab:error_sentences}
    \centering
    \resizebox{\columnwidth}{!}{\begin{tabular}{lccccc}
        \toprule
        Deviation & 0 & 1 & 2 & 4 & Vanilla \\ \midrule
        Sentences Number Satisfaction Rate & 74.50 & 68.17 & 62.50 & 60.50 & 56.33 \\
        Instruction Satisfaction Rate & 42.83 & 38.17 & 35.33 & 34.33 & 10.17 \\ \bottomrule
    \end{tabular}}
    \vspace{1em}
    
    \resizebox{\columnwidth}{!}{\begin{tabular}{lccccc}
        \toprule
        Bias & 0 & 1 & 2 & 4 & Vanilla \\ \midrule
        Sentences Number Satisfaction Rate & 74.50 & 64.67 & 58.67 & 56.50 & 56.33 \\
        Instruction Satisfaction Rate & 42.83 & 40.50 & 32.67 & 31.33 & 10.17 \\ \bottomrule
    \end{tabular}}
    
\end{table}

\subsection{Computation Time}
\label{sec:time}
We conducted experiments with 20 instructions, each containing 6 constraints, using Mistral-7B. The number of trials was set to 5, consistent with the paper's settings. The average running time is summarized below:

\begin{table}[ht]
\caption{The Average Running Time for One Sample}
\label{tab:running_time}
\centering
\resizebox{\columnwidth}{!}{
\begin{tabular}{lcccccccc}
    \toprule
    Method & Vanilla & Reflexion & U-SC & BSM & Rejection Sample & ReAct & CRITIC & DVR \\ \midrule
    Running Time (s) & 5.91 & 20.53 & 41.34 & 46.21 & 32.32 & 36.48 & 37.98 & 33.91 \\ \bottomrule
\end{tabular}}
\end{table}

As shown in Table \ref{tab:running_time}, our method does not exhibit significantly higher running time compared to other baselines. Considering the performance gains (Table \ref{tab:complex_2}), our method demonstrates a balance between efficiency and effectiveness.

\subsection{Related Work Details}
\label{sec:related}
\textbf{Instruction Following.} \textbf{(1) Evaluation:} Recent studies evaluate instruction-following capability of LLMs from various perspectives \citep{dubois2024length,zhou2023instruction,jiang2023followbench,chen2024benchmarking,zhou2023instruction,he2024can}. They evaluate LLMs' instruction-following ability by testing on length \citep{dubois2024length}, format \citep{zhou2023instruction}, semantic and topic constraints \citep{chen2024benchmarking}. Most works only test LLMs on simple instructions with only 1-2 constraints. Recently, some works test on instructions with multiple constraints \citep{he2024complex,jiang2023followbench}. They find that LLMs struggle to follow complex instructions as the number of constraints increases. Moreover, there is a big performance gap between the open-source models and the closed-source models on instruction-following. \textbf{(2) Methods:} Upon finding these problems, some works \citep{chen2023comprehensive,sun2024conifer,wang2024instructions,he2024complex} use various prompting strategies to generate instructions and responses with advanced LLMs (e.g., GPT4) and then use the generated data to fine-tune open-source LLMs. While most methods consider only a few constraints, \citet{he2024complex} focus on improving LLMs' adherence to multiple constraints. They generate complex instruction datasets by merging instructions with external constraints and iteratively refine student model responses using GPT-4 as a teacher. The student model is tuned on both intermediate modifications and final refined responses.

\textbf{LLMs Using Tools.} Tools have been extensively employed to enhance the capabilities of LLMs across various domains. For instance, retrievers are used to augment the response generation of LLMs by fetching relevant information \citep{khandelwal2019generalization,goucritic}, while search engines enhance the model's access to real-time data \cite{nakano2021webgpt}. Similarly, calculators are adopted to support math reasoning of LLMs \citep{cobbe2021training}, interpreters are used to facilitate accurate code generation \citep{chen2022program, gao2023pal} and mathematical provers help in verifying theoretical proofs \citep{jiangdraft}. CRITIC \citep{goucritic} uses external API to evaluate the toxicity score of a generated response, focusing on a single predefined task. In contrast, DVR involves preparing and selecting multiple tools, allowing for greater flexibility. Moreover, while CRITIC provides feedback as a single numerical score, DVR offers textual guidance, which has more detailed analysis for response. Another key distinction is that DVR incorporates dynamic few-shot prompting, further enhancing refinement effectiveness.



\begin{figure*}[ht]
\centering
\begin{mybox}{Refinement Example}
\textbf{Prompt:} Excuse me, could you please provide text that focuses on the music-topic? use the keywords artists. highlight at least 1 key point by wrapping it with *. For example: *highlighted key point*. The number of sentences in your response should be less than 5. be sure the letter e appears at least 4 times in your response. Make sure to include at least 3 placeholder represented by square brackets, such as [address], [name]. the total number of words in your response should be at least 23.

\textbf{Response:} 

<<Music Unites>> From festivals to online communities, music continues to bridge cultural and geographical gaps like never before. *Electronic music has especially seen a surge in popularity among the younger generation*. Talented artists from all over the world have been coming together to create beautiful melodies and harmonies. \textcolor{red}{[name] is one such artist who has been making waves in the industry with his eclectic beats. His music can be found on various platforms, including [address] and [website].} \textcolor{red}{Many emerging artists, including [another name], are also experimenting with different genres.}

\textbf{Unsatisfied Constraint:} The number of sentences in your response should be less than 5.

\textbf{Feedback:} The response contains 6 sentences, which is 2 more than allowed. Please remove at least 2 sentences.

\textbf{Refined Response:}  <<Music Unites>> From festivals to online communities, music continues to bridge cultural and geographical gaps like never before. *Electronic music has especially seen a surge in popularity among the younger generation*. Talented artists from all over the world have been coming together to create beautiful melodies and harmonies. [name] is one such artist who has been making waves in the industry with his eclectic beats \textcolor{blue}{and} his music can be found on various platforms, including [address] and [website].
\end{mybox}
\caption{The refinement process example}
\label{fig:example}
\end{figure*}

\begin{figure*}[ht]
\centering
\begin{mybox}{Decomposition Prompt}
You are an advanced assistant specializing in identifying and listing output constraints from provided instructions. The instructions typically include a task related to generating content on a specific topic and one (or multiple) format constraint(s). Your goal is to focus only on extracting and listing all the format constraints required for the output, ignoring the content-related task.

Instruction:

Generate a few lines of text that touch on the topic of tv. Put your entire answer in JSON format...

Format Constraints:

\#1. Put your entire answer in JSON format.

\#2. The word 'show' should not appear in your response.

... (more examples)

Instruction:

\{current instruction\}

Format Constraints:
\end{mybox}

\begin{mybox}{Tool Selection Prompt}
You will be given a list of constraints. Each constraint belongs to a specific category. Your task is to recognize and categorize each constraint. Only output the category from the following options:

postscript, placeholder, include keyword, exclude keyword, letter frequency, keyword frequency, sentence count constraint, word count constraint, *** separator, bullet points, fixed responses, highlighted, JSON format, title format, quoted response, end phrase, no commas, all capital letters, all lowercase, capital word frequency, language restriction

Please ensure to categorize each constraint accurately according to its description. There is definitely a valid category option for each constraint. Here are examples:

Prompt: Make sure to include the word 'mutations'.

Category: include keyword

...(more examples)

Prompt: \{Current Prompt\}

Category:

\end{mybox}

\begin{mybox}{Refinement Prompt}
You are an AI assistant responsible for refining a given response. Given a prompt, its response, and the analysis of the response, your task is to modify the response according to the analysis.

\#Prompt: I'm looking for text that explores arts or culture, can you assist? There should be no commas in your reply......

\#Original Response: Art has the power to bring people together and transcend cultural boundaries. It can evoke emotions and spark conversations that might not be possible through other means. *At the [address] museum, ......

\#It does not satisfy the constraint: There should be no commas in your reply.

\#Analysis: The response contains 4 comma(s). Here are the detected commas: (museum, visitors) (installations, each)... Please remove all commas.

\#Modified Response: Art has the power to bring people together and transcend ...(more examples)

\#Prompt: current prompt

\#Original Response: current response

\#It does not satisfy the constraint: current unsatisfied constraint

\#Analysis: current feedback

\#Modified Response: 
\end{mybox}

\caption{The prompts used in DVR}
\label{fig:prompts}
\end{figure*}

\begin{figure*}[ht]
\centering

\begin{mybox}{Word Counting}
This example is obtained through GPT4-o with zero-shot. It demonstrates that reliable tools can be easily created. The details are as follows:
\begin{verbatim}
def feedback(response, max_words=None, min_words=None):
    # Count the number of words in the response
    word_count = len(response.split())
    
    # Check for maximum word constraint
    if max_words is not None and word_count > max_words:
    return f"Response failed because it has {word_count} words, 
    exceeding the maximum allowed limit of {max_words} words."
    
    # Check for minimum word constraint
    if min_words is not None and word_count < min_words:
    return f"Response failed because it has only {word_count} 
    words, fewer than the minimum required {min_words} words."
    
    # If all constraints are satisfied
    return True
\end{verbatim}
 \end{mybox}

\begin{mybox}{Lowercase Validation}
This example validates whether a given text is entirely in lowercase. If any word contains uppercase letters, it provides feedback on which words need correction. The implementation is as follows:
\begin{verbatim}
class LowercaseLetter:
    def __init__(self):
        pass

    def feed_back(self, value):
        # Split the input string into words
        words = value.split()
        
        # Find words that are not fully in lowercase
        upper_case_words = [word for word in words if 
        any(char.isupper() for char in word)]

        if value.islower():
            return True
        else:
            return f"The response contains words that are not in 
            all lowercase letters: {', '.join(upper_case_words)}.
            Please lowercase all of them."
\end{verbatim}
 \end{mybox}
  \caption{The tool examples}
  \label{fig:tools}
\end{figure*}

\end{document}